\pdfoutput=1

\documentclass[11pt]{article}

\usepackage[]{emnlp2021}

\usepackage{times}
\usepackage{latexsym}

\usepackage[T1]{fontenc}
\usepackage[utf8]{inputenc}

\usepackage{microtype} 

\usepackage{booktabs}       
\usepackage{makecell}
\usepackage{arydshln}
\usepackage{url}
\usepackage{hyperref}
\usepackage{amsmath}
\usepackage{verbatim} 

\DeclareMathOperator*{\argmax}{arg\,max}

\usepackage{graphicx}
\graphicspath{{images/}}

\title{Subword Mapping and Anchoring across Languages} 

\author{Giorgos Vernikos$^{1,2}$ \\[4pt] $^{1}$HEIG-VD \\  Yverdon-les-Bains, Switzerland \\
\texttt{georgios.vernikos} \\ \texttt{@heig-vd.ch}
         \And
         Andrei Popescu-Belis$^{1,2}$ \\[4pt] $^{2}$EPFL School of Engineering \\ Lausanne, Switzerland \\
\texttt{andrei.popescu-belis} \\ \texttt{@heig-vd.ch}\\}

\begin{document}
\maketitle

\begin{abstract}
State-of-the-art multilingual systems rely on shared vocabularies that sufficiently cover all considered languages.  To this end, a simple and frequently used approach makes use of subword vocabularies constructed jointly over several languages. We hypothesize that such vocabularies are suboptimal due to \textit{false positives} (identical subwords with different meanings across languages) and \textit{false negatives} (different subwords with similar meanings).  To address these issues, we propose Subword Mapping and Anchoring across Languages (SMALA), a method to construct bilingual subword vocabularies. SMALA extracts subword alignments using an unsupervised state-of-the-art mapping technique and uses them to create cross-lingual anchors based on subword similarities.  We demonstrate the benefits of SMALA for cross-lingual natural language inference (XNLI), where it improves zero-shot transfer to an unseen language without task-specific data, but only by sharing subword embeddings.  Moreover, in neural machine translation, we show that joint subword vocabularies obtained with SMALA lead to higher BLEU scores on sentences that contain many false positives and false negatives.

\end{abstract}

\section{Introduction}
\label{sec:intro}

NLP systems that operate in more than one language have been proven effective in tasks such as cross-lingual natural language understanding and machine translation \cite{devlin-etal-2019-bert, conneau-etal-2020-unsupervised, aharoni-etal-2019-massively}. The performance of such systems is strongly connected to their use of an input space that can sufficiently represent all the considered languages \cite{sennrich-etal-2016-neural, wu-dredze-2019-beto, conneau-etal-2020-unsupervised}. Conceptually, an effective cross-lingual input space should exploit latent similarities between languages.

State-of-the-art multilingual systems take advantage of cross-lingual similarities in their input spaces through the use of a shared vocabulary of subwords. This vocabulary is learned on the concatenation of multilingual training corpora, using heuristic subword segmentation algorithms \cite{sennrich-etal-2016-neural, WordPiece, Kudo2018-xx}, which handle the open vocabulary problem by identifying tokens at multiple granularity levels, based on character n-gram frequencies.  Therefore, the embeddings of subwords that appear in several languages act as anchors between these languages and, thus, provide implicit cross-lingual information that leads to improved performance~\cite{xlm, pires-etal-2019-multilingual, conneau-etal-2020-emerging}.

Cross-lingual transfer in joint subword models may be limited by false positives, i.e.\ identical subwords with different meanings in two languages, a phenomenon also known as `oversharing'  \citep{Wang2020Cross-lingual, dhar-bisazza-2021-understanding}.  Moreover, they do not benefit from false negatives, i.e.\ different subwords with identical meanings. Examples of false positives are: \textit{die}, a definite article in German and a verb in English; \textit{also}, meaning `so' or `therefore' in German, not `as well' as in English; or \textit{fast}, which in German means `almost', not `quick'.  Examples of false negatives are \textit{and} and \textit{und}, \textit{very} and \textit{sehr}, \textit{people} and \textit{Menschen} -- all pairs being near synonyms that could benefit from a unique embedding rather than two.  A unique embedding would not constrain the models to always represent or translate them in the same way, as representations are highly contextualized.

In this paper, we address the problem of false positives and negatives by employing \textit{subword similarity to create cross-lingual anchors}.  Specifically, using cross-lingual mapping, we determine subword alignments for a set of subwords, and then share their representations. 
In this way, we relax the requirements for isomorphism and common scripts between languages on which previous studies rely.  We demonstrate that this can improve both cross-lingual transfer of language models and machine translation (MT).  Our contributions are the following:
\begin{enumerate} \setlength{\itemsep}{0pt}
\item We propose a method for subword mapping and anchoring across two languages (SMALA), with no constraints on the availability of parallel data or  the similarity of scripts (Section~\ref{sec:approach}).
\item We show how SMALA can be used to extend an existing monolingual vocabulary and facilitate cross-lingual transfer of a pre-trained language model to an unseen language under a limited parameter budget (Section~\ref{sec:lm-transfer}).
\item We demonstrate experimentally the benefits of SMALA for cross-language natural language inference (XNLI) (Section~\ref{sec:experiments-with-xnli}).
\item We demonstrate how SMALA can be used to build a shared vocabulary for MT, and bring experimental evidence of its benefits (Section~\ref{sec:smala-for-mt}).
\end{enumerate}
We release our code online\footnote{\url{https://github.com/GeorgeVern/smala}}.


\section{Related Work}
\label{sec:related-work}

\textbf{Cross-lingual representations}.  A large body of work has attempted to harness the similarities of languages via cross-lingual word embeddings, i.e.\ continuous word vectors that can represent multiple languages in a shared vector space. A first approach to obtain these embeddings is offline mapping of pre-trained monolingual embeddings, where the mapping can be learned using supervision in the form of lexicons \citep{DBLP:journals/corr/MikolovLS13,xing-etal-2015-normalized,joulin-etal-2018-loss}, or by leveraging weak supervision in the form of identical seed words \citep{artetxe-etal-2017-learning,sogaard-etal-2018-limitations}, or in an unsupervised way \citep{artetxe-etal-2018-robust,lample2018word}.  A second approach to obtain cross-lingual embeddings is joint training from scratch, by combining monolingual language modeling objectives with a cross-lingual objective --  with either strong, or weak, or no supervision \citep[see respectively][]{luong-etal-2015-bilingual,duong-etal-2016-learning,lample-etal-2018-phrase}. 

Despite their success, both approaches have certain limitations. On the one hand, alignment methods assume that the monolingual embedding spaces have comparable structures, i.e., that they are isomorphic to a certain extent. However, this assumption has been challenged, especially for etymologically distant languages, but also for related ones 
\citep{sogaard-etal-2018-limitations,patra-etal-2019-bilingual,ormazabal-etal-2019-analyzing}.  Unsupervised joint training, on the other hand, relies on the assumption that identical tokens carry the same information across languages, which is not always true. 

To address the limitations of alignment and joint training (the isomorphism assumption and requirement for common script), combinations of the two methods have been proposed.  \citet{Wang2020Cross-lingual} jointly train embeddings on concatenated monolingual corpora and then ``unshare'' identical words across languages, reallocating the overshared word embeddings and subsequently aligning them. \citet{ormazabal2020offline} find word alignments that are used as anchors to create cross-lingual representations with a modified version of Skip-gram \citep{skip-gram}.  Our approach shares a similar motivation, but instead of directly creating cross-lingual representations, we shape the input space (i.e.\ the vocabulary) of multilingual systems in a way that facilitates cross-lingual transfer. 

\noindent\textbf{Subword vocabularies}.
Recently, multilingual language models have superseded cross-lingual word embeddings, not only because they produce contextualized representations, but also because they can handle the open vocabulary problem through the use of subwords as tokens \citep{sennrich-etal-2016-neural, WordPiece,Kudo2018-xx}.  
Multilingual subword vocabularies are simply obtained by learning the subwords on the concatenation of all used languages.  Since each subword is assigned to a unique embedding, identical subwords that appear in several languages serve as anchors between languages, providing implicit cross-lingual information  \citep{wu-dredze-2019-beto,pires-etal-2019-multilingual,conneau-etal-2020-emerging}.  Parameter sharing across languages make subword models particularly suitable for multilingual NLP and machine translation.


The number of shared tokens in multilingual vocabularies highly depends on the similarities of script between languages.  When this is not the case, transliteration can be applied \citep{nguyen-chiang-2017-transfer,muller2020unseen,Amrhein_2020}.
In addition, shared subword vocabularies often produce inconsistent segmentations across languages that can hurt cross-lingual transfer. Regularization techniques that introduce randomness in the tokenization process \citep{Kudo2018-xx,provilkov2020bpe} can partially address this problem, or consistency between the different segmentations can be otherwise enforced \citep{MVR}. Still, there is no guarantee that shared (sub)words have identical meanings (false positives are not excluded) and, conversely, subwords with identical meanings but different spellings (false negatives) are missed.


\noindent\textbf{Cross-lingual LM transfer}. The success of pretrained monolingual and multilingual language models raises the question of whether these models can be transferred to unseen languages.
To transfer such a model, it is mostly necessary to add language-specific parameters in the form of a subword embedding layer, which can be learned from scratch \citep{artetxe-etal-2020-cross,devries2020good}.  Alternatively, offline mapping can be used to initialize the new embedding layer, for faster convergence and improved zero-shot performance \citep{tran2020english}.  Another option, which reduces the computational cost of this transfer but assumes similarity of scripts, is to leverage common subwords between languages \citep{chronopoulou-etal-2020-reusing,wang-etal-2020-extending}.  Our proposal combines the two approaches without the requirement for a common script.

Recent work has shown that cross-lingual transfer can still be achieved in the absence of anchors (i.e.\ subwords shared between languages), although the existence of anchors contributes to performance gains \citep{artetxe-etal-2020-cross, conneau-etal-2020-emerging, aji-etal-2020-neural}. Specifically, \citet{conneau-etal-2020-emerging} have shown that performance increases with the number of available anchors.  However, these studies do not discuss the quality of anchors, or how they can be obtained, which is the main focus of our work.


\section{SMALA: Subword Mapping and Anchoring across Languages}
\label{sec:approach}

Our motivation is to create cross-lingual vocabularies that are parameter-efficient and exploit the similarity of concepts between different languages. We propose a method for Subword Mapping and Anchoring across Languages (SMALA), which combines the powerful initialization of mapping methods with the anchoring properties of joint training, while attempting to alleviate the limitations of both methods.  We first learn subwords separately for each language and then train the corresponding embeddings.  We then apply a mapping method to obtain similarity scores between the embeddings, which we use to extract alignments between subwords of the two languages.  We finally tie the parameters of the aligned subwords to create anchors during training. We describe hereafter in detail the two main components of our approach.

\subsection{Subword Mapping}
\label{sec:alignment-of-subwords}
As a first step, we aim to find subwords that have similar meanings or functions (morphological or syntactic) between different languages, i.e.\ to extract subword alignments.  To this end, we first learn separate subword vocabularies for each language from monolingual data using one of the existing subword segmentation algorithms (specified below for each series of experiments).  Since we argue against using identical subwords as anchors between languages, we employ a distributional method to find the alignments: we obtain subword representations for each language from monolingual data from FastText embeddings \citep{bojanowski-etal-2017-enriching}\footnote{The use of subword co-occurrence and PCA appeared to underperform with respect to FastText.} and then align them using a state-of-the-art unsupervised alignment approach, VecMap \citep{artetxe-etal-2018-robust}. 

Our method can also exploit parallel data, when it is available.  In this case, we tokenize both sides of the bitext with language-specific subwords and then use FastAlign \citep{dyer-etal-2013-simple} to estimate the alignment, similar to \citet{tran2020english}. Implementation details can be found in Appendix~\ref{appendix:impl_details}.


\subsection{Anchoring of Similar Subwords} 
\label{sec:anchoring-of-similar-subwords}

After the mapping step, we apply cosine similarity\footnote{We also experimented with CSLS retrieval \cite{lample2018word} but it produced more alignments of lower quality.} to compute a similarity matrix $S$: each of its coefficients $S_{i,j}$ is the cosine similarity between the embeddings of the $i^\mathrm{th}$ subword of language $\mathcal{L}_1$ and of the $j^\mathrm{th}$ subword of language $\mathcal{L}_2$. 

We use the similarity matrix $S$ to identify alignments between subwords in a completely unsupervised way.  We extract the aligned subword alignments using the \textit{Argmax} method of  \citet{jalili-sabet-etal-2020-simalign}, as follows. A subword $w^{L_1}_i$ from the $\mathcal{L}_1$ vocabulary is aligned to a subword $w^{L_2}_j$ from the $\mathcal{L}_2$ vocabulary, if and only if $w^{L_2}_j$ is the most similar subword to $w^{L_1}_i$ 
and vice versa:
\begin{equation} \label{eq:mutual-argmax}
i = \argmax_l(S_{l,j}) \ \text{and} \ j = \argmax_l(S_{i,l}) 
\end{equation}
Each pair of subwords that satisfies the above consistency condition forms an alignment, to which we assign a score: the average similarity $(S_{i,j} + S_{j, i})/2$.  This will be used as a threshold to select a subset of all alignments.  We thus obtain a dictionary $D$ of aligned subwords that will function as anchors between languages during training, by tying their embeddings.

The above definition implies that the aligned subwords are translations of one another. Although this might seem quite limiting, the same issue arises for joint vocabulary construction, with the difference being the criterion according to which we choose to share subwords. We argue that our similarity is a more expressive criterion than the raw surface form.  Our approach does not rely on the surface form for cross-lingual anchors and additionally removes the requirement for a common script.  Furthermore, it prevents sharing subwords that are identical but differ in meaning (false positives) and allows sharing subwords that are spelled differently but are close to synonyms (false negatives).  The (sub)words aligned by our method may or not be identical, as long as they satisfy Equation~\ref{eq:mutual-argmax}. 


\section{Language\,Model\,Transfer\,with\,SMALA} 
\label{sec:lm-transfer}

For the first set of experiments, we attempt to transfer a pretrained Language Model (LM) from one language ($\mathcal{L}_1$) to another language ($\mathcal{L}_2$), by leveraging the linguistic knowledge that was implicitly encoded in $\mathcal{L}_1$'s embedding layer.  Following previous work \citep{artetxe-etal-2020-cross,tran2020english}, we create an embedding layer for $\mathcal{L}_2$ and initialize it by sharing parameters using SMALA.  In this way, we aim to reduce the computational budget of cross-lingual transfer via parameter sharing without sacrificing performance, but removing the need for a common script and the pitfalls of false positives and false negatives.

We transfer the model following the same steps as \citet{tran2020english}.  We start from a pretrained LM that we continue training on masked language modeling (MLM) using monolingual data from both the original and the target languages ($\mathcal{L}_1$ and $\mathcal{L}_2$). The bilingual model has two \textit{separate embedding layers}, one for $\mathcal{L}_1$ and one for $\mathcal{L}_2$, while the rest of the encoder is common to $\mathcal{L}_1$ and $\mathcal{L}_2$. Each language-specific embedding layer is used both as the first and last layer (tied embeddings).  During this training phase, we keep including monolingual data from $\mathcal{L}_1$ to avoid degradation in performance in the original language and maximize cross-lingual transfer \cite{pires-etal-2019-multilingual, conneau-etal-2020-emerging}.  We update the weights of the whole model during this phase, since updating only the embeddings would not significantly
reduce computation time (due to the need to calculate all activations for backpropagation) and has actually a negative impact on performance, as we observed in our initial experiments.  At this stage, the transferred model could be used for any cross-lingual natural language understanding task \citep{pmlr-v119-hu20b} or for unsupervised machine translation \citep{xlm,chronopoulou-etal-2020-reusing, doi:10.1162/tacl/a00343}.

In a second stage, we fine-tune the model for XNLI \citep{conneau-etal-2018-xnli} on labeled data in $\mathcal{L}_1$ (English), using $\mathcal{L}_1$ embeddings and freezing the embedding layer. Finally, we zero-shot transfer the model to $\mathcal{L}_2$ data by simply changing the language-specific embedding layer.


\section{Experiments with XNLI}
\label{sec:experiments-with-xnli}

\subsection{Models} 
\label{sec:models} 
We compare several models in our experiments on cross-lingual natural language inference (textual entailment) with the XNLI dataset \citep{conneau-etal-2018-xnli}. We note that all models, with the exception of mBERT, follow the pipeline from the previous section to transfer the pretrained LM to a new language. The only difference between these models is the way the new embedding layer is created.

\noindent\textbf{\textsc{joint}}. A system that employs parameter sharing based on surface form, that is, the union of the two language-specific vocabularies, similar to joint tokenization. The embeddings for the tokens that are not shared with the original embedding layer are initialized randomly. 

This model allows for a comparison between anchoring identical vs.\ semantically similar subwords identified by SMALA, as an inductive bias for cross-lingual vocabularies.  Although this is not exactly the same as joint tokenization, previous works have suggested that performance is similar \citep{aji-etal-2020-neural, conneau-etal-2020-emerging} and that a language-specific embedding layer and tokenizer can have a positive impact on performance \citep{rust-etal-2021-good, pfeiffer2020unks}. 

\noindent\textbf{\textsc{ours}}. Our approach (SMALA) leverages similarity to find alignments between subwords.  The parameters of the subwords are then tied, as explained above.  Our system is directly comparable to \textsc{joint}, since we only use monolingual data to find the alignments, and the non-aligned subwords are randomly initialized.

\noindent\textbf{\textsc{ours+align}}. Random initialization of the non-aligned subwords requires more computation to reach convergence \citep{artetxe-etal-2020-cross} and/or can lead to subpar performance\footnote{In our experiments, even a random alignment produced better results than random initialization.}  \citep{tran2020english, aji-etal-2020-neural}. Therefore, we also propose a system which initializes the non-aligned subwords using the similarity matrix $S$ from which we calculated the subword alignments. Following \citet{tran2020english}, we use \textit{sparsemax} \citep{pmlr-v48-martins16} to initialize the non-shared $\mathcal{L}_2$ subwords as a sparse weighted sum of $\mathcal{L}_1$ subwords. We experiment with either monolingual or parallel data to learn the similarity matrix $S$ in this case.

\noindent\textbf{\textsc{ramen}}. \textsc{ramen} \citep{tran2020english} leverages alignments learned from either monolingual or parallel data to initialize the $\mathcal{L}_2$ subword embeddings. Unlike our approach, for monolingual data, common words are used to initialize a supervised word alignment method \citep{joulin-etal-2018-loss}, and then the word alignment is transferred to subwords using several approximations.  In contrast to our method, \textsc{ramen} does not employ any parameter sharing but trains a full embedding layer for $\mathcal{L}_2$. 

\noindent\textbf{m\textsc{BERT}}. For comparison, we use multilingual BERT \citep{devlin-etal-2019-bert} in the same zero-shot cross-lingual transfer setting.  However, results are not strictly comparable to the above models, since mBERT has a larger shared vocabulary, hence more parameters (178M compared to 133M for \textsc{ramen}) and is trained for more steps. We include mBERT in our experiments as a reference for high-performing multilingual models.

\subsection{Data and Settings}

For XNLI experiments, we select five target languages that vary in terms of language family, typology and script: Spanish (Es), German (De), Greek (El), Russian (Ru) and Arabic (Ar). We obtain monolingual corpora from the Wikipedia of each language using WikiExtractor\footnote{\href{https://github.com/attardi/wikiextractor}{https://github.com/attardi/wikiextractor}}. We use these corpora for MLM training, similar to \citet{devlin-etal-2019-bert}, and to extract subword alignments using SMALA.  When parallel data is used, we either use Europarl~\cite{koehn-etal-2007-moses} or the United Nations Parallel Corpus~\cite{ziemski-etal-2016-united}. We use the same amount of parallel data for each pair and we subsample the data, if needed. Both monolingual and parallel data are lowercased and tokenized with the Moses tokenizer~\cite{koehn-etal-2007-moses}.

For our implementation we use Hugging Face’s Transformers library~\cite{HF} and for \textsc{ramen} we use the public implementation from the author. 
We choose \textsc{BERT-base} (110M parameters) as our pretrained LM. We further train all bilingual models on MLM for 120k steps with a batch size of 76 and a maximum sequence length of 256. Each batch contains equal numbers of samples from both languages, similar to \citet{tran2020english}. We optimize bilingual LMs using Adam \citep{DBLP:journals/corr/KingmaB14} with bias correction, a learning rate of 5e$-$5 and linear decay. 

\begin{table*}[ht!]
    \centering
    \begin{tabular}{lcccccc}
        \Xhline{2\arrayrulewidth} 
         Method &Data &Es &De &El &Ru &Ar \\ \hline
         \textsc{joint} &mono &$70.0 \pm 0.2$ &$64.4 \pm 0.8$ &$61.2 \pm 0.9$ &$56.2 \pm 1.1$ &$45.8 \pm 0.4$ \\ 
         \textsc{ours} &mono &$\mathbf{74.2} \pm 0.4$ &$\mathbf{70.6} \pm 0.1$ &$\mathbf{70.0} \pm 0.7$ &$\mathbf{65.4} \pm 0.9$ &$\mathbf{62.3} \pm 0.4$ \\ \hline
         \textsc{ours+align} &mono &$76.5 \pm 0.4$ &$72.8 \pm 0.5$ &$72.9 \pm 0.5$ &$70.2 \pm 0.6$ &$67.0 \pm 0.4$ \\ 
         \textsc{ours+align} &para &$\mathbf{77.1} \pm 0.8$ &$\mathbf{74.1} \pm 0.5$ &$\mathbf{75.1} \pm 0.7$ &$\mathbf{71.9} \pm 0.4$ &$\mathbf{67.8} \pm 0.8$ \\ \hline 
         \textsc{ramen} &mono &$76.5 \pm 0.6$ &$72.5 \pm 0.8$ &$72.5 \pm 0.8$ &$68.6 \pm 0.7$ &$66.1 \pm 0.8$ \\ 
         \textsc{ramen} &para &$\mathbf{77.3} \pm 0.6$ &$\mathbf{74.1} \pm 0.9$ &$\mathbf{74.5} \pm 0.6$ &$\mathbf{71.6} \pm 0.8$ &$\mathbf{68.6} \pm 0.6$ \\ 
         mBERT &mono &$74.9 \pm 0.4$ &$71.3 \pm 0.6$ &$66.6 \pm 1.2$ &$68.7 \pm 1.1$ &$64.7 \pm 0.6$ \\ \Xhline{2\arrayrulewidth}
    \end{tabular}
    \caption{\label{tab:xnli_results}
Zero-shot classification scores (accuracy) on XNLI: mean and standard deviation over 5 runs, when either monolingual or parallel corpora are used for alignment (or token matching for \textsc{joint}). Systems in the first 4 rows use parameter sharing, while those in rows 5-6 train a full embedding layer. Moreover, rows 1-2 only share subwords, while rows 3-4 also use alignment for initialization.  The best model in each subgroup is in \textbf{bold}.}
\end{table*}

We fine-tune the adapted bilingual LMs on the MultiNLI dataset \citep{N18-1101} in \textit{English}, using a batch size of 32 and a maximum sequence length of 256.  We also use Adam with a learning rate of 2e$-$5, a linear warm up schedule over the 10\% initial steps, bias correction and linear decay. We fine-tune each model for five epochs and evaluate five times per epoch, as suggested by \citet{dodge2020finetuning}. We select the best model based on validation loss.

We evaluate on the test data for $\mathcal{L}_2$ from the XNLI dataset \citep{conneau-etal-2018-xnli}, with no specific training for $\mathcal{L}_2$ (zero-shot).  As in the robust evaluation scheme for zero-shot cross-lingual transfer used by \citet{wu-dredze-2020-explicit}, we report mean and variance over the systems resulting from five different runs of the fine-tuning stage, with the same hyper-parameters but different seeds. We did not perform any exhaustive hyper-parameter search for this task, and use the exact same settings for all model variants and languages.

For each target language, we learn a new subword vocabulary using the WordPiece\footnote{As implemented at: \href{https://huggingface.co/docs/tokenizers/python/latest/}{https://huggingface.co/docs/tokenizers /python/latest/}.} algorithm \citep{WordPiece}.  The bilingual models contain two language-specific embedding layers corresponding to these vocabularies.\footnote{Following \citet{tran2020english}, we initialize special tokens ([CLS], [SEP], [MASK], [PAD] and [UNK]) with their pretrained representations, in all methods except mBERT.}  For \textsc{ramen}, which does not share parameters, the size of the $\mathcal{L}_2$ embedding layer is the same as the original one. 
For methods that employ sharing (\textsc{ours} and \textsc{joint}), the parameters of the shared subwords are tied, reducing the size of the new embedding layer. Table~\ref{tab:parameter_efficiency} presents the percentage of the $\mathcal{L}_2$ embeddings that are shared with $\mathcal{L}_1$ for all methods.

\subsection{Results on XNLI} 

We present the results of our experiments on XNLI in Table~\ref{tab:xnli_results}.  Our approach is significantly better than sharing based on surface form (\textsc{ours} vs.\ \textsc{joint}), and the improvement increases with the distance of $\mathcal{L}_2$ from English (for Greek, Russian and Arabic).  This can be attributed to the erroneous sharing of non-informative subwords (e.g. letters and English words) in the \textsc{joint} model. 

Our approach is more parameter-efficient than \textsc{joint}, as shown in Table~\ref{tab:parameter_efficiency}, as it enables 
the sharing of a larger number of embeddings, especially for distant languages.  Therefore, despite the smaller number of parameters, results are significantly improved.  Moreover, the results also demonstrate the applicability of our approach to languages with different scripts. 

Among methods that do not make use of parallel data (rows 1-3 and 5 in Table~\ref{tab:xnli_results}), we notice a significant gap between the performance of anchoring based on surface form (\textsc{joint}) and training a full embedding layer, without sharing, initialized by alignment (\textsc{ramen} with mono). Our approach can sufficiently bridge this gap, with a smaller number of parameters, demonstrating the importance of the choice of anchors in cross-lingual vocabularies.

\begin{table}[t]
        \centering
    \resizebox{\columnwidth}{!}{%
    \begin{tabular}{lcccccc}
    \Xhline{2\arrayrulewidth}
         Method &Data &Es &De &El &Ru &Ar \\ \hline
         \textsc{joint} &mono &$26\%$ &$25\%$ &$11\%$ &$9\%$  &$10\%$ \\ 
         \textsc{ours} &mono &$44\%$ &$37\%$ &$33\%$ &$31\%$ &$30\%$ \\ \textsc{ours} &para &$32\%$ &$26\%$ &$21\%$ &$21\%$ &$15\%$ \\ 
    \Xhline{2\arrayrulewidth}
    \end{tabular}
    }
    \caption{\label{tab:parameter_efficiency} Percentage of $\mathcal{L}_2$  embeddings that are shared with $\mathcal{L}_1$ (English) for each system and language.}

\end{table}


Among methods that use alignment (rows 3-6), our approach with additional alignment of the non-shared subwords (\textsc{ours+align}) performs on par or better than \textsc{ramen}. This trend is consistent across the use of monolingual and parallel data for the alignment. In the latter case, the alignment is learned with the same method and data in both systems. Our higher score 
supports
our claim that better anchoring can lead to more parameter-efficient vocabularies without sacrificing performance.

Finally, in Table~\ref{tab:xnli_results}, we observe that all methods that employ alignment outperform mBERT. In some cases, even our approach without alignment performs comparably (Es, De) or even better (El) than mBERT. These results show that our method -- which transfers a monolingual LM to an unseen language with minimal computation demands -- is a competitive alternative to using an off-the-shelf multilingual model. This is particularly useful when the considered language is not modeled well (e.g.\ Greek) or not covered at all by the multilingual model.

\begin{table*}[t] 
\begin{center}
\begin{tabular}{lcccccccc}
\Xhline{2\arrayrulewidth}
Languages & \multicolumn{2}{c}{En-Ru} & \multicolumn{2}{c}{En-De } &
  \multicolumn{2}{c}{En-Ro } & \multicolumn{2}{c}{En-Ar } \\
Data & \multicolumn{2}{c}{25M} & \multicolumn{2}{c}{5.85M} &
  \multicolumn{2}{c}{612k} & \multicolumn{2}{c}{239k} \\
&  $\leftarrow$ & $\rightarrow$ & $\leftarrow$ & $\rightarrow$ &
$\leftarrow$ & $\rightarrow$ & $\leftarrow$ & $\rightarrow$ \\
\hline
\textsc{joint} & 30.0 & 26.1 & 32.1 & 27.1 & 30.9 & 23.2 & 29.0 & 11.8 \\
\textsc{ours} & 30.2 & \textbf{26.6} & 32.1 & 27.0 & 30.8 & 23.3 & 28.8 & 12.2 \\
\Xhline{2\arrayrulewidth}
\end{tabular}
\caption{\label{tab:MT_results_1col} BLEU scores of baseline and our system for machine translation.  Language pairs are ordered by decreasing size of training data (numbers of sentences). \textbf{Bold} indicates statistical significance ($p<0.05$).}
\end{center}
\end{table*} 

\section{Experiments with Machine Translation} 
\label{sec:smala-for-mt}

In the second set of experiments, we apply SMALA to MT by leveraging subword alignments to create shared bilingual vocabularies from scratch, 
instead of joint subword vocabularies learned on concatenated source and target corpora. 

\subsection{Applying SMALA to MT}

The majority of current Transformer-based MT systems \citep{NIPS2017_3f5ee243} share the vocabulary and the corresponding embedding layer between the encoder and the decoder of a sequence-to-sequence architecture.  To apply SMALA to MT, 
instead of jointly learning the subwords on the concatenated corpora, 
we learn separate subword vocabularies for each language, and then merge them into a joint one.  We use SMALA to extract alignments from the available parallel data of each language pair, and use aligned pairs as unique subwords (shared entries), serving as anchors in the shared embedding layer.  These anchors play the same role as identical subwords in joint vocabularies, and thus address the problem of false negatives.  Conversely, identical subwords that are not aligned with SMALA remain two distinct language-specific entries, thus addressing the problem of false positives.

To create a subword vocabulary of a given size $n$ using SMALA, we first learn two monolingual vocabularies of size $m > n$, one for the source and one for the target language. Then, we select a number of alignments $\alpha$ with the highest similarity scores, as defined in Section~\ref{sec:anchoring-of-similar-subwords}, with $\alpha = 2m - n$.  This ensures that, when the two vocabularies are joined and the $\alpha$ pairs of anchors are merged, the size of the resulting vocabulary is $n$.

\subsection{Data, Tools and Settings}

We choose four language pairs that represent different levels of data availability and language relatedness, and run experiments in both directions: Russian, German, Romanian and Arabic, to and from English.
Training and test data comes from WMT17\footnote{\href{http://statmt.org/wmt17/translation-task.html}{http://statmt.org/wmt17/translation-task.html}} for En-Ru and En-De, WMT16\footnote{\href{http://statmt.org/wmt16/translation-task.html}{http://statmt.org/wmt16/translation-task.html}} for En-Ro, and IWSLT17\footnote{TED talks from: \href{https://wit3.fbk.eu/}{https://wit3.fbk.eu/}} for En-Ar.
We tokenize the data using the Unigram LM model \citep{Kudo2018-xx} as implemented in SentencePiece\footnote{\href{https://github.com/google/sentencepiece}{https://github.com/google/sentencepiece}}.
We choose the size of the shared subword vocabulary based on the size of the data, following \citet{Kudo2018-xx}: 32k for high-resource pairs (En-Ru and En-De) and 16k for medium and low-resource pairs (En-Ro and En-Ar). We tokenize data using the Moses Tokenizer~\cite{koehn-etal-2007-moses}. 
We report BLEU scores \citep{papineni-etal-2002-bleu} obtained with SacreBLEU \citep{post-2018-call} on detokenized text.\footnote{Signature: BLEU+c.mixed+\#.1+.exp+tok.13a+v.1.5.1}

We train OpenNMT-py~\cite{klein-etal-2017-opennmt} for a maximum of 100k steps on high-resource pairs and 40k steps on medium or low-resource ones.  Our base model is Transformer-Base ($L$=6, $H$=512) \citep{NIPS2017_3f5ee243} with the same regularization and optimization procedures.  We use a batch size of 4k tokens and evaluate every 5k steps.  We select the best model based on validation loss.  Final translations are generated with a beam width of five. 

\subsection{Results}
\label{sec:results-nmt}

We present the results for our method and the baseline in Table~\ref{tab:MT_results_1col}.  Our method yields comparable results to the baseline across all conditions of data availability and language relatedness. This demonstrates the viability of SMALA as an alternative for the creation of shared bilingual vocabularies. We observe a slight increase in performance in distant language pairs (En-Ru and En-Ar), which could be explained by the difference in scripts.  Indeed, joint tokenization (baseline system) is not able to identify anchors when the script is not shared between languages, resorting to a small number of shared subwords that are mostly uninformative, often due to the presence of English words in the other language.  In this case, the anchors found by SMALA (subword pairs corresponding to false negatives in the baseline) help to improve the joint vocabulary.

Comparing the results of Tables~\ref{tab:xnli_results} and \ref{tab:MT_results_1col} we see that our approach does not equally improve results in both settings. We attribute this difference to the amount of supervision available in MT in the form of bitext, and to the strong contextual constraints from the decoder. Although false positives and negatives are present in both scenarios, the availability of parallel data for training forces NMT models to disambiguate these subwords based on context in both languages at the same time. 

\section{Analysis}
In this section we attempt to quantify the effect of false positives and false negatives on each of the tasks.

\subsection{Ablation Study on XNLI}\label{sec:ablation_xnli}
We begin with a model that creates cross-lingual anchors based on surface form (\textsc{joint}) and we address either false positives only (\textsc{$-$fp}) or false negatives only (\textsc{$-$fn}) among shared subwords. In the latter case, if a subword is both a false positive and a false negative, then we treat it as a false negative -- e.g., \textit{also} in English should be not aligned with \textit{also} in German but with \textit{auch}.
We follow the pipeline of Section~\ref{sec:lm-transfer} and present the results in XNLI in Table~\ref{tab:only_fp_fn}.

\begin{table}[ht]
        \centering
    \resizebox{\columnwidth}{!}{%
    \begin{tabular}{lcccccc}
    \Xhline{2\arrayrulewidth}
         Method  &Es &De &El &Ru &Ar \\ \hline
         \textsc{joint} &$70.0$ &$64.4$ &$61.2$ &$56.2$  &$45.8$ \\ 
         \textsc{$-$fp} &$68.5$ &$61.7$ &$62.6$ &$53.6$ &$44.8$ \\ 
         \textsc{$-$fn} &$74.3$ &$70.0$ &$70.2$ &$65.8$ &$63.1$ \\ 
         \textsc{ours ($-$fp$-$fn)}
         &$74.2$ &$70.6$ &$70.0$ &$65.4$ &$62.3$ \\
    \Xhline{2\arrayrulewidth}
    \end{tabular}
    }
    \caption{\label{tab:only_fp_fn} Effect of removing false positives or false negatives in XNLI (accuracy).}
\end{table}


We observe that by only removing false positives (\textsc{$-$fp}) performance drops compared to \textsc{joint}. This can be attributed to the ability of the model to disambiguate false positives in the presence of context. But this could also be due to a limitation of our method to identify false positives with high precision especially (sub)words that have more than one sense. Conversely, the problem of false negatives seems to be the most important and by addressing it (\textsc{$-$fn}) results improve significantly over \textsc{joint}. The similar performance of \textsc{$-$fn} and \textsc{ours} may be due to the removal of certain false positives along with many false negatives (see also Appendix~\ref{appendix:alignments_smala}).

\subsection{False Positives and Negatives in MT}

In order to quantify the effect of false positives and false negatives in MT, we compare the performance of joint tokenization with SMALA for cases where the presence of such subwords is significant. Table~\ref{tab:FP_FN_results} presents BLEU scores for sentences that contain a high percentage of false positives and/or negatives (more than 50\% of the tokens) in the source side, along with the number of sentences in this case. BLEU scores for percentages between 0\% and 60\% are represented graphically in the Appendix, Figure~\ref{fig:FP_FN_plot}.

\begin{table}[ht]
\begin{center}
\resizebox{\columnwidth}{!}{%
\begin{tabular}{lcccccccc}
\Xhline{2\arrayrulewidth}

Languages 
& 
\multicolumn{2}{c}{En-Ru} &
\multicolumn{2}{c}{En-De} &
\multicolumn{2}{c}{En-Ro} &
\multicolumn{2}{c}{En-Ar}
\\

& 
$\leftarrow$ & $\rightarrow$ &
$\leftarrow$ & $\rightarrow$ &
$\leftarrow$ & $\rightarrow$ &
$\leftarrow$ & $\rightarrow$ 
\\

\hline

Sentences &
49 & 2225  & 
1674 & 2216 & 
1249 & 1295 &
141 & 866 
\\

\hline

\textsc{joint} &
39.2 &
27.6 &
33.1 &
27.0 &
31.6 &
24.6 &
37.8 &
16.2 
\\

\textsc{ours} &
42.2 &
28.0 &
33.0 &
27.0 &
32.0 &
24.8 &
40.4 &
16.6
\\

$\Delta$ &
+3.0 &
+0.4 &
-0.1 &
0.0 &
+0.4 &
+0.2 &
+2.6 &
+0.3
\\

\Xhline{2\arrayrulewidth}
\end{tabular} 
}
\caption{\label{tab:FP_FN_results} 
BLEU scores for sentences where 50\% of tokens are false positives and/or false negatives.  The number of selected sentences (out of a total of 3,000) is given for each translation direction.}
\end{center}
\end{table}

The results of Table~\ref{tab:FP_FN_results} show improved performance of our method over the baseline, confirming our original intuition regarding false positives and negatives.  Despite the fact that MT models with joint tokenization use context to disambiguate false positives -- as it can help to also disambiguate polysemous words to a certain extent \citep{rios-gonzales-etal-2017-improving,pu-etal-2018-integrating} -- when their number increases performance tends to drop compared to SMALA. The gap in performance between \textsc{joint} and \textsc{ours} (using SMALA) is bigger for pairs that do not have shared scripts (En-Ru and En-Ar) which is a possible indication of the impact of false negatives, despite the smaller sample sizes.  Overall, the results of Tables~\ref{tab:MT_results_1col} and \ref{tab:FP_FN_results} demonstrate that our approach is competitive with joint tokenization in most cases and superior in challenging cases with multiple false positives and negatives. 

\subsection{Cross-lingual Word Representations}
In order to validate our claim that SMALA facilitates cross-lingual transfer, we perform an intrinsic evaluation of the obtained representations. We compare the quality of representations created using SMALA vs.\ joint tokenization
for Bilingual Lexicon Induction (BLI), a standard evaluation task for cross-lingual word embedding methods. Specifically, we compare the performance of the bilingual models from the first setting (see Section~\ref{sec:lm-transfer}) after the bilingual MLM training step, but  before the XNLI fine-tuning. We do not include methods that use alignment to initialize the embedding layer (for these results see Appendix \ref{appendix:bli}), in order to isolate the effect of anchors.

We follow the setting of \citet{vulic-etal-2020-probing} to compute word-level representations.
We encode each word in isolation using the model, in the form [CLS] \textit{word} [SEP]. We extract the representations from the embedding layer excluding representations of special tokens. If a word is split into more than one subword, we average the obtained representations. We perform this operation for every word of the test set for both languages. We retrieve word translation using \textit{Cross-Domain Similarity Local Scaling} (CSLS) with $K$=10 number of neighbours \citep{lample2018word}.

\begin{figure}[ht]
    \centering
    \includegraphics[height=5cm, keepaspectratio]{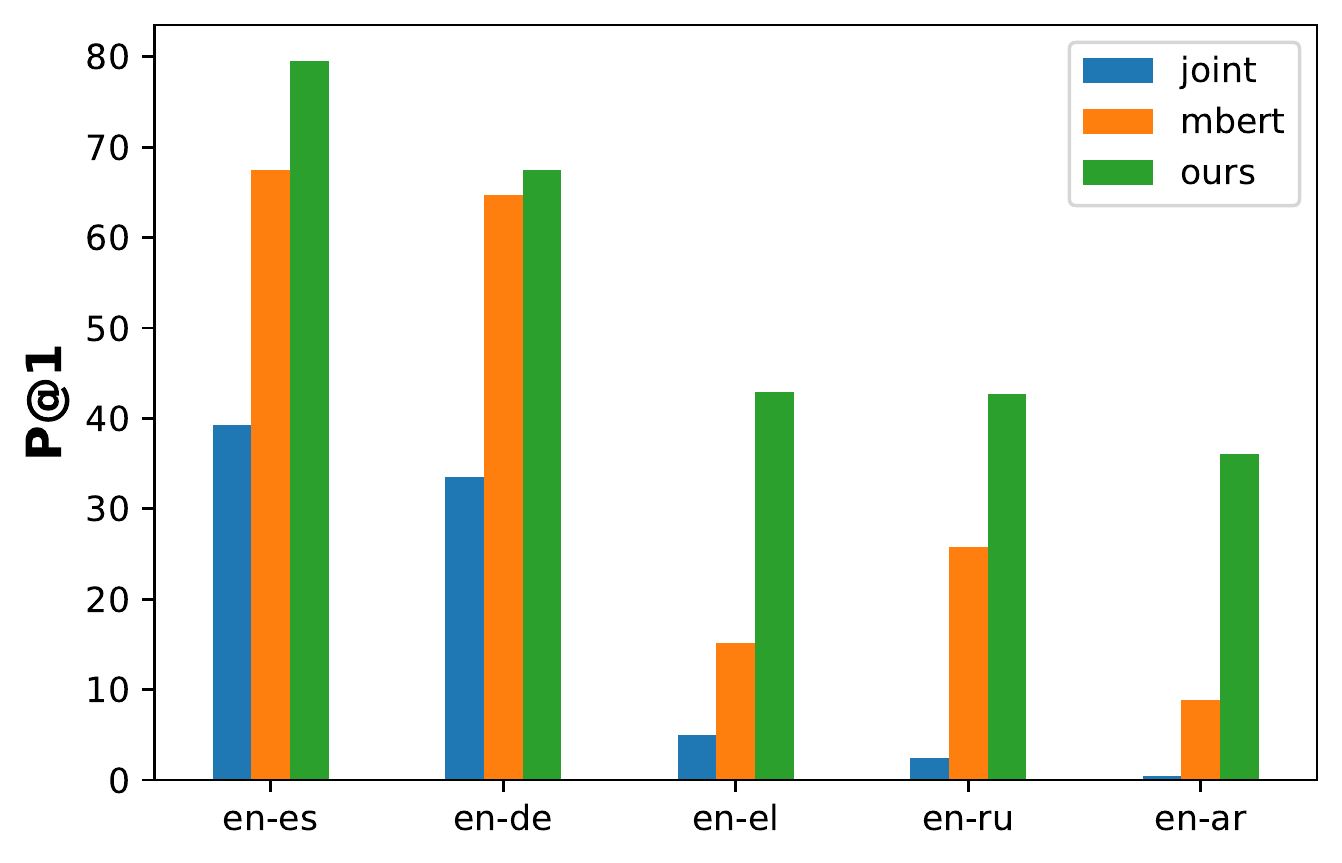}
    \caption{Precision@1 results for the BLI task.}
    \label{fig:BLI_results}
\end{figure}

Our results on the MUSE benchmark~\cite{lample2018word}, a bilingual dictionary induction dataset, are presented in Figure~\ref{fig:BLI_results}, using
precision at 1 scores (P@1), following standard practices. 
We observe that by using SMALA to create cross-lingual anchors (\textsc{ours}) we can greatly improve performance on BLI compared to methods that use identical subwords (\textsc{joint} and mBERT). 
Figure~\ref{fig:BLI_results} also shows that the performance of \textsc{joint} and mBERT significantly decreases as the two languages are more distant and their vocabulary does not have considerable overlap, which points at the limitations of joint tokenization and especially false negatives which are the most frequent in this case. 

\begin{figure}[ht]
    \centering
    \includegraphics[height=5cm, keepaspectratio]{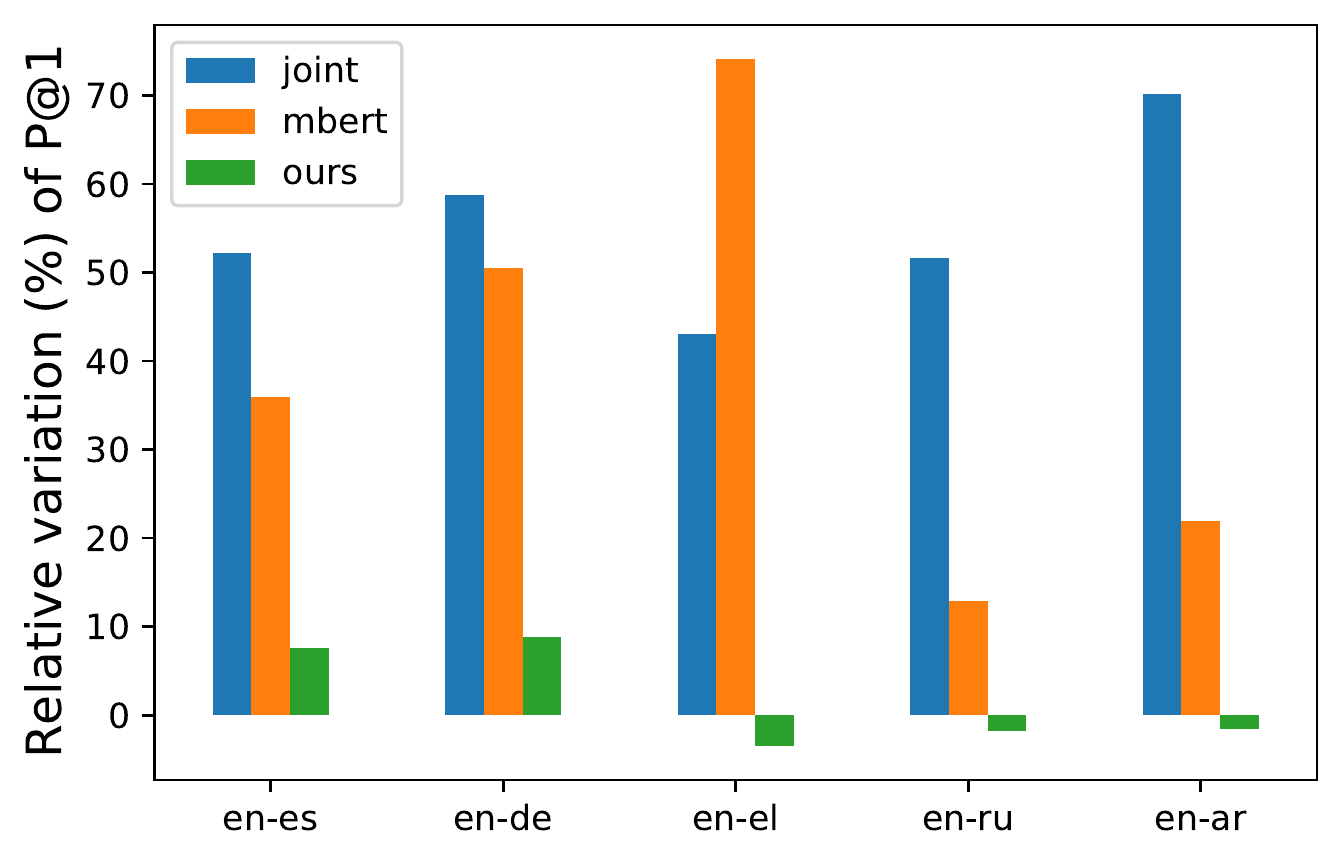}
    \caption{Precision@1 difference on BLI when test pairs of same surface form are removed.}
    \label{fig:BLI_drop}
\end{figure}

Similar to \citet{Wang2020Cross-lingual}, we also evaluate on words that are not shared, by removing test pairs with the same surface form (e.g.\ (\textit{epic}, \textit{epic}) as a test pair for en-es) and present the difference in performance in Figure~\ref{fig:BLI_drop}. We find that the performance of \textsc{joint} and mBERT
decreases significantly, unlike \textsc{ours}. For languages with different scripts 
(en-el, en-ru and en-ar) 
the performance of our approach even increases in this scenario due to the fact that our system is able identify and not retrieve false positives. This confirms our intuition that the use of surface form to create cross-lingual anchors leads to poorly aligned cross-lingual representations for the non-shared subwords.

\section{Conclusion}
In this work we introduced SMALA, a novel approach to construct shared subword vocabularies that leverages similarity instead of identical subwords to create anchors. We demonstrate that our approach outperforms current methods for joint construction of multilingual subword vocabularies in cases where there is no cross-lingual signal, apart from the anchors. When cross-lingual supervision is available, our approach performs comparably to the baseline, 
while showing improved performance in cases with numerous false positive and false negatives. 

In future work, we aim to extend our method to more than two languages. We also intend to explore the effectiveness of SMALA for closely related languages and compare SMALA to other approaches, such as those using transliteration. In addition, we aim to apply SMALA to settings of varying cross-lingual supervision levels, such as unsupervised MT.


\section*{Acknowledgments} 
We are grateful for their support to
the Swiss National
Science Foundation through grant n.\ 175693 for the DOMAT project: ``On-demand Knowledge for
Document-level Machine Translation'' and to  Armasuisse for the FamilyMT project.

\bibliographystyle{acl_natbib}
\bibliography{references}

\clearpage
\appendix

\section{Appendix}
\label{sec:appendix}

\subsection{SMALA Implementation Details} 
\label{appendix:impl_details}

To train subword embeddings we use FastText~\cite{bojanowski-etal-2017-enriching} with dimension 1,024. Other than that, we use the default parameters, i.e.\ a window size of 5 and 10 negative examples. For the mapping of the embedding we use the unsupervised version of VecMap~\cite{artetxe-etal-2018-robust} with default hyperparameters. Indeed, we argue against identical subwords as cross-lingual anchors, and the unsupervised version takes advantage of similarity distributions of equivalent words in a way that matches our intuition. If parallel data is available, we use FastAlign~\cite{dyer-etal-2013-simple} with default hyperparameters. Our approach is not bound to these specific tools and can benefit from future research in the fields of (sub)word representation and (supervised or unsupervised) alignment.

\subsection{Alignments Produced by SMALA} 
\label{appendix:alignments_smala}

The number of alignments of SMALA depends on the language relatedness and the amount of monolingual and multilingual data. In Table~\ref{tab:num_shared_LM} we present the number of subwords that are shared between languages for the first set of experiments (XNLI). We note that the maximum number of shared subwords is $30,522$ (the number of $\mathcal{L}_1$ subwords). 

\begin{table}[h]
    \centering
    \resizebox{\columnwidth}{!}{%
    \begin{tabular}{l|c|c|c|c|c|c}
    \Xhline{2\arrayrulewidth}
         Method &Data &Es &De &El &Ru &Ar \\ \hline
         \textsc{joint} &mono &$7,936$ &$7,554$ &$3,395$ &$2,836$  &$2,909$ \\ 
         \textsc{ours} &mono &$13,466$ &$11,269$ &$10,120$ &$9,334$ &$9,245$ \\ 
         \textsc{ours} &para &$9,708$ &$7,945$ &$6,491$ &$6,265$ &$4,590$\\ \Xhline{2\arrayrulewidth} 
         \textsc{ramen} &* &$0$ &$0$ &$0$ &$0$ &$0$ \\ 
    \Xhline{2\arrayrulewidth}
    \end{tabular}
    }
        \caption{Number of $\mathcal{L}_2$ subword embeddings that are shared with $\mathcal{L}_1$ for each system and language.}
    \label{tab:num_shared_LM}
\end{table}

In Table~\ref{tab:anchors_abl} we present the number of shared subwords (anchors) for the ablation experiments of Section~\ref{sec:ablation_xnli}. The number of false positives identified by SMALA can be computed as the difference between the first and the second row, e.g.\ $7,780-4,374=3,406$ for Es. The number of false negatives can computed as the difference between the fourth and the second row, e.g.\ $13,466-4,374=9,092$ for Es. The difference between 
the number of false positives and the difference between the number of anchors of \textsc{$-$fn} and \textsc{ours} reveals how many false positives are removed while removing false negatives, e.g.\ $3,406-(15,269-13,466)=1,603$ for Es.

\begin{table}[ht]
        \centering
    \resizebox{\columnwidth}{!}{%
    \begin{tabular}{l|c|c|c|c|c}
    \Xhline{2\arrayrulewidth}
         Method  &Es &De &El &Ru &Ar \\ \hline
         \textsc{joint} &$7,780$ &$7,395$ &$3,283$ &$2,685$  &$2,743$ \\ 
         \textsc{$-$fp} &$4,374$ &$3,838$ &$285$ &$286$ &$230$ \\ 
         \textsc{$-$fn} &$15,269$ &$13,189$ &$11,727$ &$10,826$ &$10,770$ \\ 
         \textsc{ours}
         &$13,466$ &$11,269$ &$10,120$ &$9,334$ &$9,245$ \\
    \Xhline{2\arrayrulewidth}
    \end{tabular}
    }
    \caption{\label{tab:only_fp_fn_subwords} Number of shared subwords in the case of only false positives or only false negatives. \textsc{ours} amounts to \textsc{$-$fp$-$fn}.}
    \label{tab:anchors_abl}
\end{table}

For MT, we choose the number of monolingual vocabularies so that the merged vocabulary is equal in size to the one produced by joint tokenization. This leads to monolingual vocabularies of size $20$k for En-De, $18.5$k for En-Ru, $10$k for En-Ro and $9$k for En-Ar.

\subsection{Scores on Validations Sets} 
Tables~\ref{tab:xnli_dev_results} and \ref{tab:MT_results_dev} present the results on the development sets for the two sets of experiments. 

\subsection{Model Training Details} 
The amount of shared subwords of Table~\ref{tab:num_shared_LM} translates to fewer parameters in the first setting. For Spanish (Es), for example, the number of added parameters (on top of the 110M parameters of pretrained BERT) for \textsc{ours} with mono is  $(30,522-13,466)\times768$ compared to $30,522\times768$ for \textsc{ramen}, where 768 is the dimension of the token embeddings.

We train the bilingual LMs of Section~\ref{sec:models} on two GeForce GTX 1080 Ti GPUs for approximately 55 hours. We then fine-tune our models on one GPU for 8 hours, except for mBERT where we use two due to the increased number of parameters.

For MT, the Transformer model for the high-resource pairs has 60.6M parameters and for the medium and low-resource pairs 52.4M, due to the difference in vocabulary size. For these experiments, we train the high-resource models on the same two GPUs for 50 hours and the medium/low-resource ones for 20 hours.

\subsection{Additional Results on BLI}
\label{appendix:bli}

\begin{figure*}[ht]
    \centering
    \includegraphics[height=7cm, keepaspectratio]{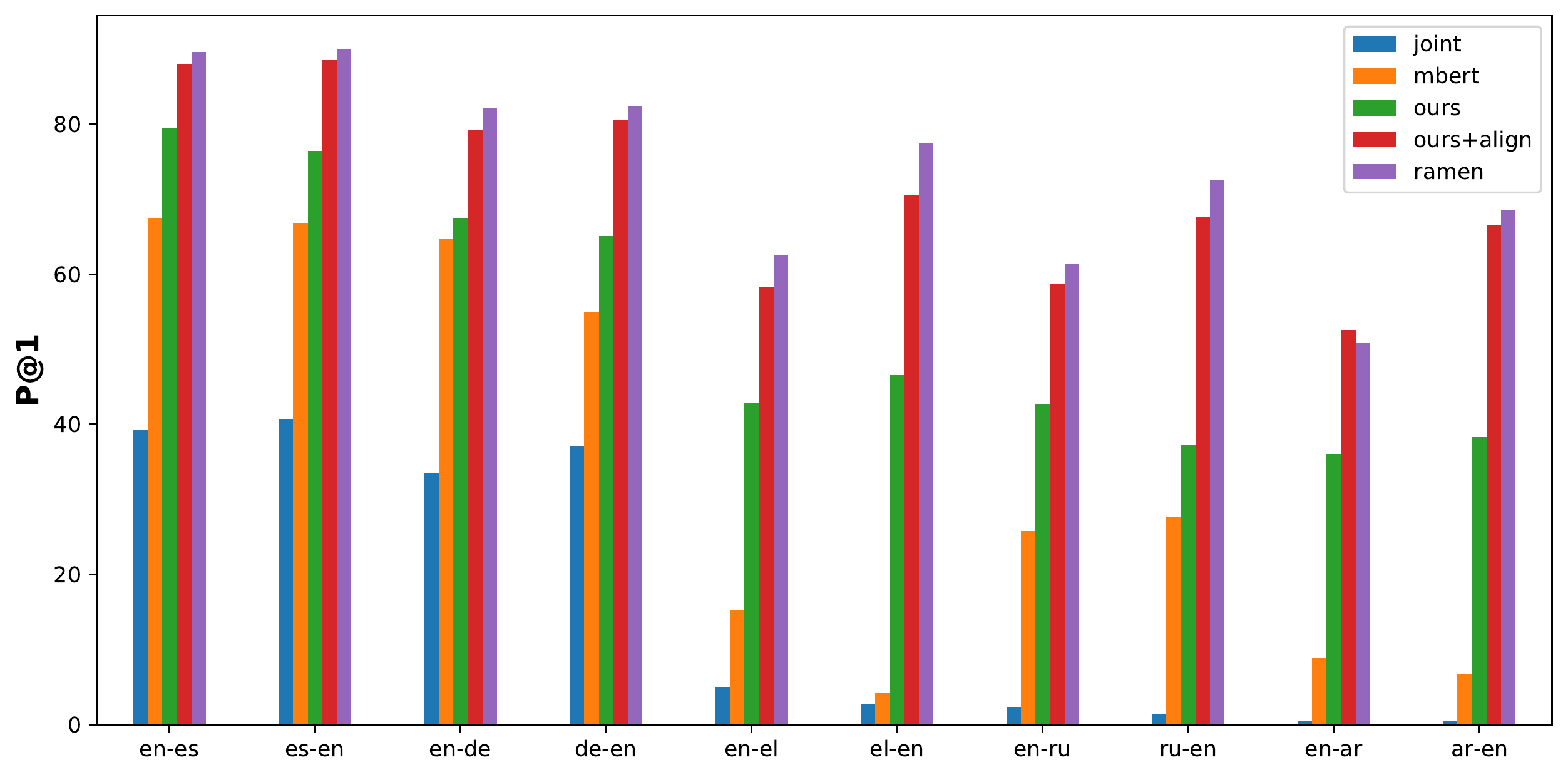}
    \caption{Precision@1 results for the BLI task.}
    \label{fig:BLI_more_results}
\end{figure*}

Figure~\ref{fig:BLI_more_results} presents results on BLI for all methods and both directions. We also include models that use alignment for the initialization of their embeddings (i.e \textsc{ours+align} and \textsc{ramen}), but only compare methods that use monolingual data.  The initialization of non-shared subwords further improves results, which is expected since it provides a cross-lingual signal for all subword representations.  

Furthermore, \textsc{ramen} slightly outperforms \textsc{ours+align}, which could be attributed to the larger number of parameters. Another reason could

\begin{table*}[ht]
\begin{tabular}{p{\columnwidth}p{\columnwidth}}
  be the inductive bias of \textsc{smala}, which leads to retrieval of the aligned target (sub)word for a given source (sub)word, ignoring other possible translations. Although this might hurt cross-lingual rep- & resentations if context is absent (i.e.\ subword embeddings), our results show that it improves performance for zero-shot cross-lingual transfer. \\
\end{tabular}
\end{table*}

\begin{figure*}[ht]
    \centering
    \includegraphics[height=17cm, keepaspectratio]{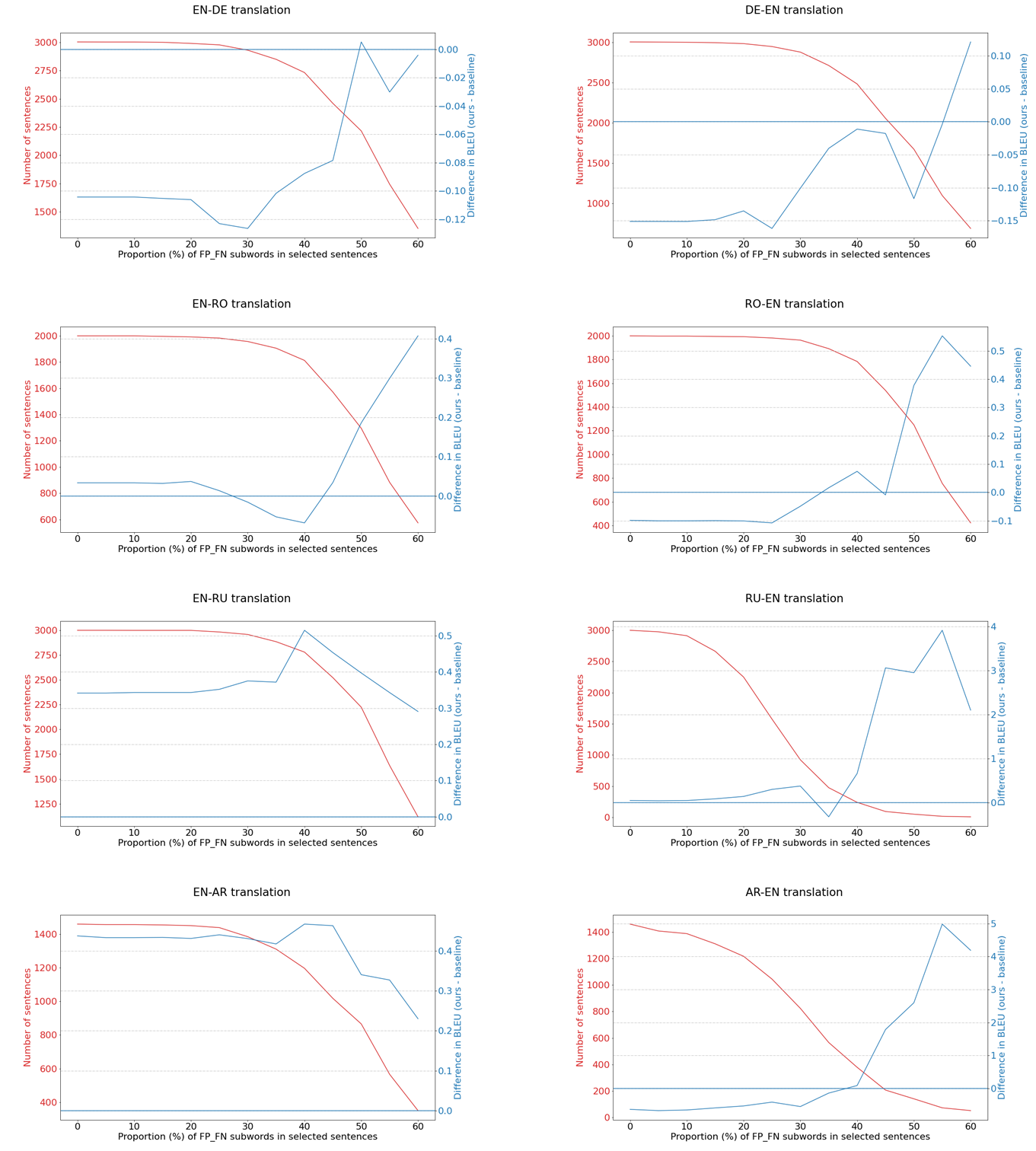}
    \caption{BLEU scores for sentences that contain a high percentage of false positives and/or false negatives.}
    \label{fig:FP_FN_plot}
\end{figure*}


\begin{table*}[ht]
    \centering
    \begin{tabular}{lcccccc}
        \Xhline{2\arrayrulewidth} 
         Method &Data &Es &De &El &Ru &Ar \\ \hline
         \textsc{joint} &mono &$70.2\pm1.2$ &$64.5\pm1.2$ &$61.0\pm0.9$ &$56.3\pm1.2$ &$45.5\pm0.5$ \\ 
         \textsc{ours} &mono &$74.3\pm0.5$ &$69.6\pm0.6$ &$68.6\pm0.9$ &$65.5\pm1.3$ &$62.6\pm0.3$ \\ \hdashline
         \textsc{ours+align} &mono &$76.5\pm0.4$ &$71.9\pm0.6$ &$71.9\pm0.5$ &$68.9\pm0.9$ &$65.8\pm0.2$ \\ 
         \textsc{ours+align} &para &$76.5\pm0.8$ &$73.7\pm0.6$ & $75.3\pm0.7$ &$70.3\pm0.8$ &$66.9\pm0.9$ \\ \Xhline{2\arrayrulewidth} 
         \textsc{ramen} &mono &$75.5\pm0.8$ &$72.0\pm1.3$ &$72.2\pm0.4$ &$67.7\pm0.9$ &$64.9\pm0.8$ \\ 
         \textsc{ramen} &para &$76.9\pm0.8$ &$73.9\pm1.2$ &$74.9\pm0.9$ &$69.7\pm0.7$ &$68.1\pm1.3$ \\ \hline
         mBERT &mono &$74.6\pm0.6$ &$72.1\pm0.7$ &$66.3\pm1.2$ &$68.5\pm1.0$ &$62.9\pm0.8$ \\ \Xhline{2\arrayrulewidth}
    \end{tabular}
    \caption{\label{tab:xnli_dev_results}
Zero-shot classification scores on XNLI dev set (accuracy): mean and standard deviation over five runs are reported. Results follow the same format as those in Table~\ref{tab:xnli_results}.}
\end{table*}

\begin{table*}[ht] 
\begin{center}
\begin{tabular}{lcccccccc}
\Xhline{2\arrayrulewidth}
& \multicolumn{2}{c}{En-Ru} & \multicolumn{2}{c}{En-De} &
  \multicolumn{2}{c}{En-Ro} & \multicolumn{2}{c}{En-Ar} \\
&  $\leftarrow$ & $\rightarrow$ & $\leftarrow$ & $\rightarrow$ &
$\leftarrow$ & $\rightarrow$ & $\leftarrow$ & $\rightarrow$ \\
\hline
\textsc{joint} &30.0  &27.8  &34.6  &31.7  &33.1  &26.5  &33.1  &15.5  \\
\textsc{ours} &30.2  &28.3  &34.6  &31.6  &33.0  &26.1 &31.8  &15.5 \\
\Xhline{2\arrayrulewidth}
\end{tabular}
\caption{\label{tab:MT_results_dev} BLEU scores (detokenized) of baseline and our system for machine translation on the development set.}
\end{center}
\end{table*} 

\end{document}